\documentclass[letterpaper]{article} 
\usepackage[preprint]{aaai2027}  
\nocopyright
\usepackage[hyphens]{url}  
\usepackage{graphicx} 
\usepackage{natbib}  
\usepackage{caption} 
\usepackage{algorithm}
\usepackage{algorithmic}

\usepackage{newfloat}
\usepackage{listings}
\DeclareCaptionStyle{ruled}{labelfont=normalfont,labelsep=colon,strut=off} 
\floatstyle{ruled}
\newfloat{listing}{tb}{lst}{}
\floatname{listing}{Listing}

\usepackage{booktabs}
\usepackage{amsmath,amssymb}
\usepackage{multirow}
\usepackage[table]{xcolor}
\title{Let Geometry GUIDE: Layer-wise Unrolling of Geometric Priors in Multimodal LLMs}
\author{
Chongyu Wang\textsuperscript{\rm 1,\rm 2},
Ting Huang\textsuperscript{\rm 2},
Chunyu Sun\textsuperscript{\rm 3},
Xinyu Ning\textsuperscript{\rm 4},
Di Wang\textsuperscript{\rm 1}\corresponding,
Hao Tang\textsuperscript{\rm 2}\corresponding
}

\affiliations{
\textsuperscript{\rm 1}Xi'an Jiaotong University
\quad
\textsuperscript{\rm 2}Peking University
\quad
\textsuperscript{\rm 4}InspireOmni\\
\textsuperscript{\rm 3}Institute of Automation, Chinese Academy of Sciences\\
chongyu@stu.xjtu.edu.cn,
diwang@xjtu.edu.cn,
bjdxtanghao@gmail.com
}

\usepackage{etoolbox}

\newcommand{\arxivteaser}{%
  \vspace{-1.2em}
  \begin{center}
    \includegraphics[
      width=0.96\textwidth,
      height=0.32\textheight,
      keepaspectratio
    ]{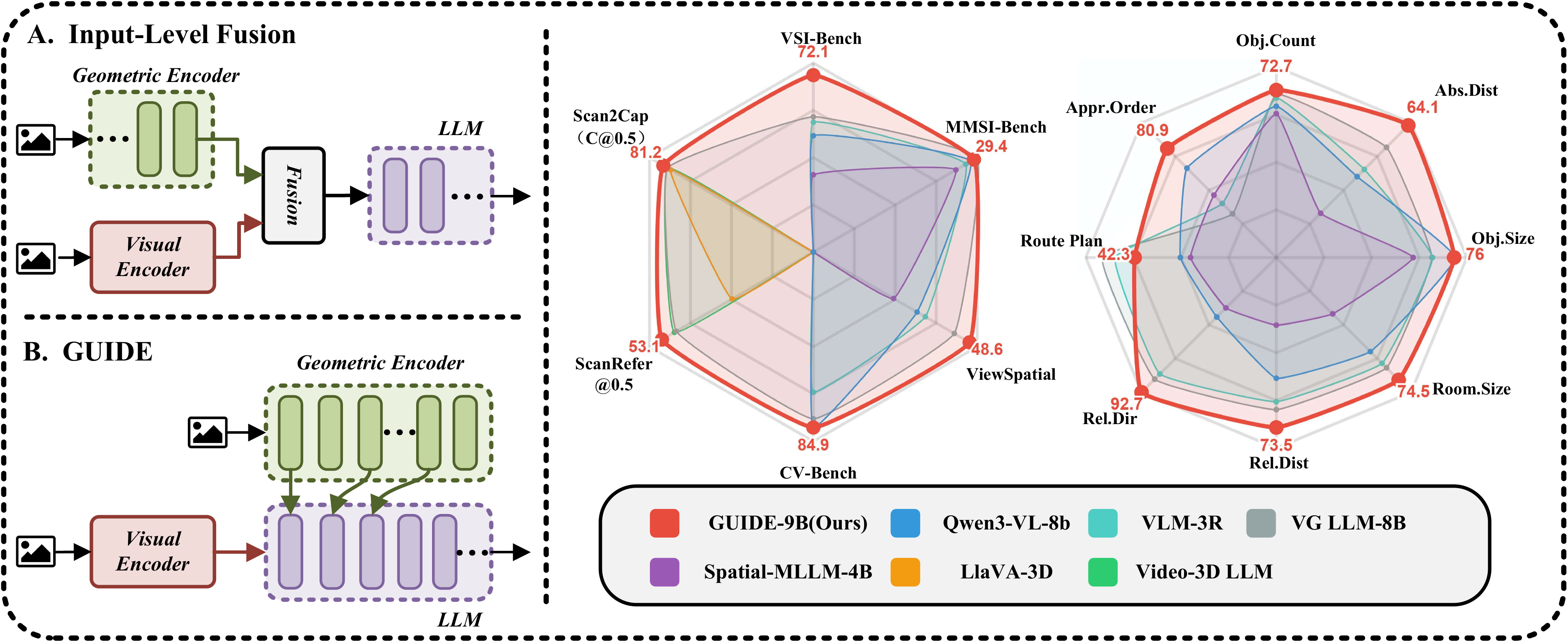}
    \captionof{figure}{
      \textbf{Overview and performance of GUIDE.}
      \textbf{Left:} Compared with one-shot input-level fusion, GUIDE
      sequentially injects multi-level geometric features into early LLM
      layers.
      \textbf{Middle:} Results across spatial understanding benchmarks.
      \textbf{Right:} Category-wise results on
      VSI-Bench~\protect\cite{vsibench}.
    }
    \label{fig:teaser}
  \end{center}
  \vspace{0.3em}
}

\makeatletter
\patchcmd{\maketitle}
  {\twocolumn[\@maketitle]}
  {\twocolumn[\@maketitle\arxivteaser]}
  {}
  {\PackageError{arxiv-teaser}
    {Failed to patch \string\maketitle}
    {Check the definition of \string\maketitle\space in aaai2027.sty.}}
\makeatother

\begin{document}

\maketitle

\begin{abstract}
Multimodal Large Language Models (MLLMs) have achieved remarkable progress in 2D visual tasks but still struggle to understand physical space in real-world visual streams. Recently, feed-forward geometric foundation models that implicitly extract geometric priors from visual inputs have offered a promising direction. However, many existing geometry-aware MLLMs extract features from a single deep encoder layer and perform one-shot fusion at the language-model input, potentially overlooking complementary information across geometric encoder layers, limiting access to fine-grained geometric cues, and hindering progressive cross-modal adaptation.
To address these limitations, we propose GUIDE (\textbf{G}eometric \textbf{U}nrolling \textbf{I}nsi\textbf{d}e MLLM \textbf{E}arly-layers), a progressive framework for integrating geometric priors. GUIDE samples and aligns multi-level features from the geometric encoder and sequentially injects them into the early layers of the MLLM. This design enables the MLLM to continuously access multi-granularity geometric cues and progressively integrate them with visual representations during layer-wise semantic abstraction. GUIDE further introduces a dual context-aware gating mechanism at both the token and layer levels, selectively regulating geometric information to reduce redundant injection and interference with pretrained representations. Extensive experiments on spatial reasoning and 3D scene understanding benchmarks, including VSI-Bench, ScanRefer, and Scan2Cap, validate the effectiveness and cross-task applicability of GUIDE. Our 5B and 9B models achieve average scores of \textbf{71.5} and \textbf{72.1} on VSI-Bench, respectively.

\end{abstract}




\section{Introduction}


Multimodal Large Language Models (MLLMs) have achieved remarkable progress in image and video understanding and open-ended visual dialogue. Yet interacting with the physical world requires reasoning beyond 2D appearance. Reliable 3D spatial perception is essential for applications such as embodied AI, autonomous driving, and augmented reality navigation, where models must reason about metric scale, instance-level occlusion relationships, and global scene layout. Despite their strong high-level semantic understanding, existing MLLMs remain limited in recovering such physical structure from monocular RGB visual streams, which are among the most accessible forms of real-world visual input. One line of work addresses this limitation by incorporating explicit 3D representations, such as point clouds, depth maps, or 3D meshes. Although effective when high-quality geometric data is available, these approaches typically rely on depth sensors or costly offline 3D reconstruction, limiting their scalability to unconstrained real-world videos.

Recently, advances in feed-forward visual geometry foundation models, such as DUSt3R \cite{dust3r} and VGGT \cite{vggt}, have provided a new pathway for extracting geometric information from RGB visual inputs. Without relying on depth sensors or pre-constructed 3D scenes, these models can estimate dense geometric representations and camera poses through feed-forward inference, thereby facilitating the development of geometry-aware MLLMs. Existing methods have explored various strategies for geometric feature fusion. For example, VLM-3R \cite{vlm3r} and Spatial-MLLM \cite{spatialmllm} employ cross-attention mechanisms, VGLLM \cite{vgllm} uses patch-wise residual addition, and SpaceMind \cite{spacemind} introduces camera tokens as modulation conditions. Despite their different implementations, many existing methods follow a similar design: they extract features from a single deep layer of the geometric encoder (typically the final or penultimate layer) and perform one-shot fusion before the features enter the LLM backbone, as illustrated in Figure~\ref{fig:teaser}A.

Although this design can effectively introduce geometric information, two aspects remain underexplored. First, representations at different layers of a geometric encoder may contain complementary information, whereas relying on a single deep-layer feature may overlook geometric cues from other layers that are relevant to fine-grained spatial understanding. Second, one-shot input-level fusion provides geometric information only at the entrance of the LLM, requiring subsequent layers to preserve and utilize all relevant cues from the initial fusion, without providing continuous access to geometric information at different granularities as multimodal representations evolve through the network. Motivated by these observations, we investigate how to sample and align multi-level features from the geometric encoder and sequentially inject them into the early layers of the MLLM, thereby supporting the progressive integration of geometric priors with multimodal representations.

To explore this design space, we propose \textbf{GUIDE} (\textbf{G}eometric \textbf{U}nrolling \textbf{I}nsi\textbf{d}e MLLM \textbf{E}arly-layers), a progressive framework for integrating geometric priors. While retaining input-level geometric anchoring, GUIDE extends the interaction between geometric information and multimodal representations into the early layers of the LLM, as illustrated in Figure~\ref{fig:teaser}B. Specifically, GUIDE uniformly samples features from multiple layers of a feed-forward geometric encoder, aligns them with the spatial grid of visual tokens, and then sequentially injects them into successive early LLM layers. Compared with designs that rely solely on a single deep-layer geometric feature and one-shot input-level fusion, this approach leverages complementary information across different layers of the geometric encoder and provides evolving multimodal hidden representations with continuous access to multi-granularity geometric cues, thereby supporting the progressive integration of geometric priors with visual-semantic representations.

While progressive layer-wise injection provides continuous access to multi-granularity geometric cues, repeatedly introducing geometric features may also accumulate redundant signals and interfere with the pretrained representations of the LLM. To regulate this process, we introduce a \textbf{dual context-aware gating mechanism} consisting of a context-aware token-level gate and a learnable layer-level global gate. The token-level gate conditions on the evolving hidden states at visual-token positions to modulate geometric information for individual visual tokens, while the layer-level gate controls the overall injection strength at each layer. Together, these gates selectively regulate geometric information across token positions and network layers, thereby reducing redundant injection and limiting interference with pretrained representations.

Our main contributions are summarized as follows:
\begin{itemize}
    \item \textbf{Progressive Geometric Integration:} We propose the GUIDE framework to break the bottlenecks of single
deep-layer feature extraction and input-level fusion. By progressively
unrolling geometric priors within the LLM, GUIDE enables the model to
continuously access multi-granularity geometric cues.
    
    

    \item \textbf{Multi-Level Alignment and Dual Context-Aware Gating:}
We align complementary multi-level geometric features and introduce
dual context-aware gating to mitigate interference from repeated
injection, enabling effective geometric integration across multiple
early LLM layers.
    \item \textbf{Comprehensive Evaluation:}
Controlled comparisons and component-wise ablations validate the
design of GUIDE. Our 5B and 9B models achieve average scores of 71.5
and 72.1 on VSI-Bench, respectively, while results on ScanRefer and
Scan2Cap demonstrate its applicability to 3D scene understanding.
    
\end{itemize}

\begin{figure*}[t]
    \centering
    \includegraphics[height=11cm, keepaspectratio]{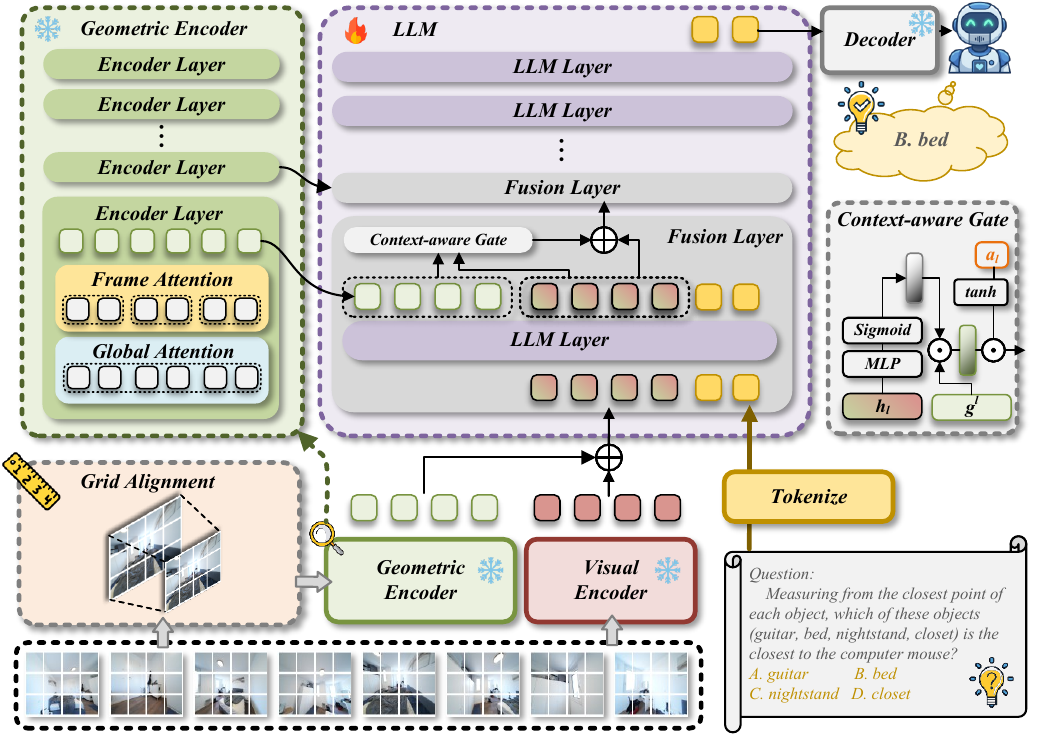}
    \caption{\textbf{Overall architecture of GUIDE}. GUIDE employs progressive unrolling to inject multi-granularity features from the 3D geometric encoder (green) into early LLM layers (purple) via a dual context-aware gate (grey), enabling the model to continuously access geometric cues at different granularities while mitigating interference from redundant geometric signals.}
    \label{fig:GUIDE_pipline}
\end{figure*}

\section{Related Work}

\noindent\subsection{Multimodal Large Language Models}

Multimodal Large Language Models (MLLMs) have evolved from contrastive vision-language pretraining, exemplified by CLIP~\cite{clip} and ALIGN~\cite{align}, to the alignment of visual encoders with LLMs through architectures such as Flamingo~\cite{flamingo} and BLIP-2~\cite{blip2}. More recent instruction-tuned models~\cite{qwen3vl,llava,internvl} have achieved strong performance in image and video understanding. Nevertheless, spatial benchmarks such as VSI-Bench~\cite{vsibench} reveal persistent limitations in metric estimation, viewpoint reasoning, and complex 3D relationships, motivating the development of MLLMs with stronger spatial representations.

\noindent\subsection{MLLMs for Spatial Intelligence}

Existing approaches to spatially capable MLLMs broadly follow two directions. The first directly incorporates explicit 3D representations, including point clouds, RGB-D scans, and BEV features~\cite{ll3da,gpt4scene,video3dllm}. Although effective for 3D grounding and reasoning, these methods depend on depth sensors or offline reconstruction, limiting their applicability to unconstrained RGB videos.

The second direction improves spatial reasoning under RGB-only inputs through data scaling, spatial instruction tuning, or specialized optimization strategies~\cite{spacer,vilasr,ray2025sat,spar,cambrians,sensenova-si}. For example, SpatialLadder~\cite{spatialladder} employs reinforcement learning to progressively enhance spatial reasoning. These approaches avoid explicit 3D inputs, but acquire spatial knowledge primarily through training supervision rather than direct geometric representations.

\noindent\subsection{Geometry-Aware MLLMs}


Feed-forward geometry models enable dense geometric information to be extracted directly from RGB inputs~\cite{dust3r,mast3r,vggt}. Building on these models, recent geometry-aware MLLMs incorporate learned geometric priors through diverse feature alignment and fusion mechanisms~\cite{vlm3r,spatialmllm,vgllm,spacemind,geoalign,gapmllm}.

A common design in this line of research extracts geometric representations from a single deep encoder layer and performs one-shot fusion before LLM processing. Although effective, this design may overlook complementary information across encoder layers and provide limited access to geometric cues as visual-token representations evolve through the LLM. Motivated by these limitations, GUIDE samples and spatially aligns multi-level geometric features and sequentially injects them into early LLM layers. A context-aware token-level gate and a learnable layer-level global gate further regulate this process across token positions and network layers, reducing redundant geometric signals and limiting interference with pretrained representations.

\section{The Proposed Method}
\subsection{Overall Architecture}

We present \textbf{GUIDE}, a framework for progressively integrating
multi-level geometric priors into an MLLM. Given an RGB image sequence
$\mathcal{I}=\{I_i\}_{i=1}^{N}$ and a text prompt $Q$, GUIDE processes
the images using a visual encoder and a parallel feed-forward geometric
encoder. Multi-level features sampled from the geometric encoder are
spatially aligned with the visual-token grid and projected into the LLM
hidden space. GUIDE retains a deep geometric feature as an input-level
anchor and sequentially injects the aligned multi-level features into
successive early LLM layers. A context-aware token-level gate and a
learnable layer-level global gate jointly regulate the injected
geometric information across visual-token positions and network layers.
This design provides evolving visual-token representations with
continuous access to multi-granularity geometric cues while reducing
redundant injection and limiting interference with pretrained
representations.

We next describe visual and geometric feature extraction and spatial
alignment, followed by progressive layer-wise injection and
token- and layer-level gating.

\subsection{Multi-Level Geometric Feature Extraction and Alignment}
\label{sec:feature_extraction}

\noindent\textbf{Visual Representation Extraction.}
Let $P_v$ denote the patch size of the visual encoder, and let
$H_v=\lfloor H/P_v \rfloor$ and $W_v=\lfloor W/P_v \rfloor$ denote
the resulting spatial grid dimensions. The visual encoder maps the
input sequence to
\begin{equation}
T_V \in
\mathbb{R}^{N \times H_v \times W_v \times C_v}.
\end{equation}
We adopt Qwen3-VL~\cite{qwen3vl} as the MLLM backbone. Its visual
merger combines each spatially adjacent $2\times2$ group of visual
tokens and projects them into the LLM hidden dimension:
\begin{equation}
T'_V =
\phi_v\bigl(\operatorname{Merge}_{2\times2}(T_V)\bigr)
\in
\mathbb{R}^{N \times \frac{H_v}{2} \times
\frac{W_v}{2} \times C_{\mathrm{llm}}},
\end{equation}
where $\phi_v$ denotes the visual projector. 
The merged visual tokens form the visual component of the LLM input
and are further augmented by the input-level geometric anchor
described below.

\noindent\textbf{Multi-Level Geometric Feature Extraction.}
GUIDE employs a pretrained feed-forward geometry model, such as
VGGT~\cite{vggt}, as a parallel geometric encoder. Because the visual
and geometric encoders may use different patch sizes, directly
processing the same image resolution produces spatial grids of
different sizes. Let $P_g$ be the patch size of the geometric encoder.
We resize each input image to
\begin{equation}
I'_i \in
\mathbb{R}^{(H_vP_g)\times(W_vP_g)\times3},
\end{equation}
such that the geometric encoder produces an $H_v\times W_v$ feature
grid. This establishes a one-to-one spatial correspondence between
visual and geometric feature locations.

Let $\mathcal{F}_{\mathrm{geo}}=\{f^k\}_{k=1}^{K}$ denote the intermediate features produced by the $K$ layers of the
cross-frame interaction module. To emphasize features formed after
sufficient cross-frame interaction, we exclude the first $\lfloor K/4 \rfloor$ layers and uniformly sample
$m$ layers from the remaining layers:
\begin{equation}
\mathcal{S}_{\mathrm{geo}}
=
\{f^{k_j}\}_{j=1}^{m},
\qquad
k_1 < k_2 < \cdots < k_m.
\end{equation}
The sampled features capture complementary geometric information
across different encoder depths.

Each sampled feature is spatially merged and projected into the LLM
hidden dimension:
\begin{equation}
g^j =
\phi_g\bigl(
\operatorname{Merge}_{2\times2}(f^{k_j})
\bigr)
\in
\mathbb{R}^{N \times \frac{H_v}{2} \times
\frac{W_v}{2} \times C_{\mathrm{llm}}},
\end{equation}
where $\phi_g$ denotes the geometric projector. The resulting feature
set $\mathcal{G}=\{g^j\}_{j=1}^{m}$ is aligned with the merged visual
tokens in both spatial resolution and channel dimension.

\subsection{Progressive Geometric Injection with Dual Context-Aware  Gating}
\label{sec:fusion_gating}

\noindent\textbf{Input-Level Geometric Anchor.}
GUIDE retains the conventional input-level fusion used by geometry-aware
MLLMs as an initial geometric anchor. Specifically, the penultimate geometric feature $f^{K-1}$ is spatially merged and projected as
\begin{equation}
g_{\mathrm{anchor}}
=
\phi_g\bigl(
\operatorname{Merge}_{2\times2}(f^{K-1})
\bigr).
\end{equation}
It is then added to the aligned visual tokens:
\begin{equation}
T_{\mathrm{anchor}}
=
T'_V + g_{\mathrm{anchor}}.
\end{equation}
The resulting visual tokens are concatenated with the text tokens to
form the initial LLM input. Unlike relying exclusively on this
one-shot fusion, GUIDE further provides early LLM layers with
sequential access to multi-level geometric features.

\noindent\textbf{Progressive Layer-Wise Injection.}
We map the ordered geometric feature set
$\mathcal{G}=\{g^j\}_{j=1}^{m}$ to the first $m$ LLM layers, with $g^l$ injected into the output hidden states of the $l$-th LLM
layer at the corresponding visual-token positions. This
mapping provides evolving visual-token representations with geometric
features extracted at different encoder depths, rather than requiring
all relevant geometric information to be preserved from the initial
fusion alone.

Repeated layer-wise injection, however, may introduce redundant
geometric signals and excessively perturb the pretrained hidden
representations. We therefore regulate each injection at both the
token and layer levels.

\noindent\textbf{Context-Aware Token-Level Gate.}
Let $h_l^V$ denote the output hidden states at the visual-token positions of
the $l$-th LLM layer. The token-level gate is conditioned on these
evolving visual hidden states:
\begin{equation}
G_{\mathrm{tok}}^l
=
\sigma\bigl(\operatorname{MLP}(h_l^V)\bigr)
\end{equation}
where $\sigma$ denotes the sigmoid function. The MLP produces a scalar
gate for each visual token, which is broadcast across the feature
channels to independently modulate the aligned geometric information.

\noindent\textbf{Layer-Level Global Gate.}
To control the overall injection strength at each layer, we introduce
a learnable scalar $\alpha_l$:
\begin{equation}
G_{\mathrm{layer}}^l
=
\tanh(\alpha_l).
\end{equation}
The gated geometric residual is then computed as
\begin{equation}
\widehat{g}^l
=
G_{\mathrm{layer}}^l
\left(
G_{\mathrm{tok}}^l \odot g^l
\right),
\end{equation}
and applied only to the visual-token positions:
\begin{equation}
h_l^{V,\mathrm{out}}
=
h_l^V + \widehat{g}^l.
\end{equation}
The token-level gate regulates geometric information at individual
spatial positions, while the layer-level gate controls its global
contribution at each network depth. Together, they provide continuous
yet controlled access to multi-granularity geometric cues across
multiple early LLM layers, while suppressing redundant signals and
mitigating interference with pretrained representations.

\begin{table*}[t]
\centering
\resizebox{\textwidth}{!}{
\begin{tabular}{l c cccc cccc}
\toprule
\multirow{2}{*}{\textbf{Models}} & \multirow{2}{*}{\textbf{Avg.}} & \multicolumn{4}{c}{\textbf{Numerical Answer}} & \multicolumn{4}{c}{\textbf{Multiple-Choice Answer}} \\
\cmidrule(lr){3-6} \cmidrule(lr){7-10}
& & \textbf{Obj. Count} & \textbf{Abs. Dist} & \textbf{Obj. Size} & \textbf{Room Size} & \textbf{Rel. Dis} & \textbf{Rel. Dir} & \textbf{Route Plan} & \textbf{Appr. Order} \\
\midrule
\rowcolor{gray!10}
\multicolumn{10}{l}{\textbf{Proprietary Models}} \\
Gemini-2.5-pro~\cite{gemini2.5} & 53.6 & 46.0 & 37.4 & 68.7 & 54.4 & 62.0 & 43.9 & 47.4 & 68.8 \\
GPT-5~\cite{gpt5} & 55.0 & 53.3 & 34.5 & 73.3 & 47.5 & 63.7 & 48.7 & \textbf{50.3} & 68.9 \\
\midrule
\rowcolor{gray!10}
\multicolumn{10}{l}{\textbf{Open-source General Models}} \\
Bagel-7B-MoT~\cite{bagel} & 31.4 & 30.1 & 29.2 & 35.5 & 25.8 & 34.9 & 41.4 & 30.4 & 24.1 \\
Qwen3-VL-8B-Instruct~\cite{qwen3vl} & 57.9 & 67.6 & 47.0 & \textbf{76.3} & 61.9 & 58.0 & 51.0 & 35.1 & 66.3 \\
InternVL3-8B~\cite{internvl3} & 42.1 & 66.1 & 34.9 & 43.6 & 47.5 & 48.0 & 39.3 & 26.3 & 31.4 \\
VILA-1.5-40B~\cite{vila1.5} & 31.2 & 22.4 & 24.8 & 48.7 & 22.7 & 40.5 & 25.7 & 31.5 & 32.9 \\
LLaVA-NeXT-Video-72B~\cite{llavanext} & 40.9 & 48.9 & 22.8 & 57.4 & 35.3 & 42.4 & 36.7 & 35.0 & 48.6 \\
\midrule
\rowcolor{gray!10}
\multicolumn{10}{l}{\textbf{Spatial Intelligence Models}} \\ 
SpatialLadder-3B~\cite{spatialladder} & 44.9 & 62.2 & 35.4 & 62.0 & 41.4 & 45.6 & 46.5 & 27.3 & 38.5 \\
Spatial-MLLM-4B~\cite{spatialmllm} & 48.4 & 65.3 & 34.8 & 63.1 & 45.1 & 41.3 & 46.2 & 33.5 & 46.3 \\
ViLaSR-7B~\cite{vilasr} & 44.6 & 58.1 & 33.9 & 61.4 & 28.9 & 45.1 & 46.5 & 29.9 & 53.2 \\
VST-7B-SFT~\cite{vst} & 60.6 & 72.0 & 44.4 & 74.3 & 68.3 & 59.7 & 55.8 & 44.9 & 65.2 \\
Cambrian-S-3B~\cite{cambrians} & 57.3 & 70.7 & 40.6 & 68.0 & 46.3 & 64.8 & 61.9 & 27.3 & 78.8 \\
Cambrian-S-7B~\cite{cambrians} & 67.5 & \textbf{73.2 }& 50.5 & 74.9 & 72.2 & 71.1 & 76.2 & 41.8 & 80.1 \\
Spa3-VLM-4B~\cite{jiang2026spa3r} &58.6 &69.0 &39.3 &70.6 &62.8 &57.9 &59.5 &36.1 &73.6 \\
SPAR-8B~\cite{spar} & 41.1 & - & - & - & - & - & - & - & - \\
VLM-3R~\cite{vlm3r} & 60.9 & 70.2 & 49.4 & 69.2 & 67.1 & 65.4  & 80.5 & 45.4 & 40.1 \\
SenseNova-SI$_{\text{Qwen3-VL-8B}}$~\cite{sensenova-si} & 64.8 & - & - & - & - & - & - & - & - \\
SenseNova-SI$_{\text{InternVL3-8B}}$~\cite{sensenova-si} & 68.8 & - & - & - & - & - & - & - & - \\
VG-LLM-4B~\cite{vgllm} & 47.3 & 66.0 & 37.8 & 55.2 & 59.2 & 44.6 & 45.6 & 33.5 & 36.4 \\
VG-LLM-8B*~\cite{vgllm} & 62.2 & 71.4 & 56.8 & 69.0 & 69.1 & 67.9 & 83.2 & 47.4 & 32.5 \\
VG-LLM-5B$_{\text{Qwen3-VL-4B}}$ & 51.7 & 67.7 & 42.5 & 64.4 & 59.5 & 49.8 & 48.5 & 33.5 & 48.2  \\
\midrule
\rowcolor{gray!10} 
\textbf{GUIDE-5B (Ours)} & 55.6& 70.0 & 43.0 & 64.3& 67.0& 58.8& 50.5&35.6 & 56.0\\
\rowcolor{gray!10} 
\textbf{GUIDE-5B$^\dagger$ (Ours)} & 71.5& 68.8 & 61.5 & 75.6& 73.2& 73.0& 88.9&48.5 & \textbf{82.7}\\
\rowcolor{gray!10}
\textbf{GUIDE-9B$^\dagger$  (Ours)} & \textbf{72.1}& 72.7 & \textbf{64.1} & 76.0& \textbf{74.5} & \textbf{73.5}& \textbf{92.7} & 42.3&80.9 \\
\bottomrule
\end{tabular}
}
\caption{Performance on VSI-Bench. We compare GUIDE with recent general-purpose and spatially enhanced models. $^\ast$ denotes
training with the S1+S2 data configuration of VG-LLM, while
$^\dagger$ denotes our scaled-data training setting.}
\label{tab:vsibench_results}
\end{table*}

\section{Experiments}


We evaluate \textbf{GUIDE} on a range of benchmarks covering spatial reasoning and 3D scene understanding. We first describe the experimental setup
and implementation details, followed by comparisons across these two
task categories. We then conduct component-wise ablations to examine
the effects of progressive multi-level injection and the dual gating
mechanism.

\noindent\subsection{Implementation and Training Details}
\label{sec:setup}

GUIDE-5B and GUIDE-9B are instantiated with Qwen3-VL-4B and
Qwen3-VL-8B~\cite{qwen3vl}, respectively, and both employ VGGT-1B
~\cite{vggt} as the feed-forward geometric encoder. Following
VG-LLM~\cite{vgllm}, we use the feature from the 23rd VGGT layer as
the input-level geometric anchor. For progressive layer-wise injection,
we extract multi-level geometric features from VGGT layers 8, 11, 14,
17, 20, and 23 and sequentially inject them into the first 6 LLM
layers. We use Adam with a global batch size of 64, a learning rate of
$1\times10^{-5}$, and a warmup ratio of 0.03. 
We train the LLM backbone, geometric projector, and gating modules,
while keeping the visual encoder, VGGT encoder, and original
multimodal connector frozen throughout training.
All experiments are
conducted on eight NVIDIA A100 80GB GPUs.
The training data and their composition for each setting are detailed in the following two subsections.

\begin{table}[htbp]
  \centering
  \small
  \renewcommand{\arraystretch}{1.1}
  \resizebox{\columnwidth}{!}{%
  \begin{tabular}{l c |c c c c}
    \toprule
    \textbf{Models} & 
    \rotatebox{45}{\textbf{Avg.}} &
    \rotatebox{45}{\textbf{VSI}} & 
    \rotatebox{45}{\textbf{MMSI}} & 
    \rotatebox{45}{\textbf{ViewSp.}} & 
    \rotatebox{45}{\textbf{CV}} \\
    \midrule
    \rowcolor{gray!10}
    \multicolumn{6}{l}{\textbf{\textit{Proprietary Models}}} \\
    Gemini-2.5-Pro~\cite{gemini2.5} & 55.9 & 53.6 & 38.0 & 46.0 & \textbf{85.9} \\
    GPT-5~\cite{gpt5} & 56.7 & 55.0 & \textbf{41.8} & 45.5 & 84.6 \\
    \midrule
    \rowcolor{gray!10}
    \multicolumn{6}{l}{\textbf{\textit{Open-Source General Models}}} \\
    Bagel-7B-MOT~\cite{bagel} & 44.9 & 31.4 & 31.0 & 41.3 & 76.0 \\
    Qwen3-VL-8B-Instruct~\cite{qwen3vl} & 52.7 & 57.9 & 28.8 & 39.0 & 85.1 \\
    InternVL3-2B~\cite{internvl3} & 42.1 & 32.9 & 26.5 & 32.5 & 76.5 \\
    InternVL3-8B~\cite{internvl3} & 47.4 & 42.1 & 28.0 & 38.6 & 81.0 \\
    \midrule
    \rowcolor{gray!10}
    \multicolumn{6}{l}{\textbf{\textit{Spatial Intelligence Models}}} \\
    SpatialLadder-3B~\cite{spatialladder} & 46.4 & 44.9 & 27.4 & 39.8 & 73.7 \\
    Spatial-MLLM-4B~\cite{spatialmllm} & -- & 48.4 & 26.1 & 34.7 & -- \\
    Cambrian-S-3B~\cite{cambrians} & 49.2 & 57.3 & 25.2 & 39.0 & 75.2 \\
    Cambrian-S-7B~\cite{cambrians} & 52.8 & 67.5 & 25.8 & 40.9 & 76.9 \\
    VLM-3R-7B~\cite{vlm3r} & 50.3 & 60.9 & 27.9 & 40.5 & 71.8 \\
    VG-LLM-4B~\cite{vgllm} & 49.7 & 47.3 & 28.0 & 42.5 & 81.0 \\
    VG-LLM-8B*~\cite{vgllm} & 54.9 & 62.2 & 30.0 & 45.8 & 81.7 \\
    \midrule
    \rowcolor{gray!10}
    GUIDE-5B (Ours) & 54.5 & 55.6 & 30.2 & 47.3 & 84.7 \\
    \rowcolor{gray!10}
    GUIDE-5B$^\dagger$  (Ours)  & 58.3 & 71.5 & 29.6 & 47.7 & 84.5 \\
    \rowcolor{gray!10}
    GUIDE-9B$^\dagger$  (Ours)  & \textbf{58.8} & \textbf{72.1} & 29.4 & \textbf{48.6} & 84.9 \\
    \bottomrule
  \end{tabular}%
  }
  \caption{Comparison of different models on VSI-Bench~\cite{vsibench}, MMSI-Bench~\cite{mmsibench}, ViewSpatial~\cite{viewspatial}, and CV-Bench~\cite{cvbench}.}
  \label{tab:benchmark_results}
\end{table}

\noindent\subsection{Spatial Reasoning}
\label{sec:spatial_reasoning}

\noindent\textbf{Benchmarks and Evaluation Metrics.}
We use VSI-Bench~\cite{vsibench} as the primary benchmark for spatial
reasoning. It covers a broad spectrum of multimodal spatial geometric question-answering scenarios, ranging from relative direction judgment to absolute distance estimation. To evaluate
cross-benchmark generalization, we additionally report results on
MMSI-Bench~\cite{mmsibench}, ViewSpatial~\cite{viewspatial}, and
CV-Bench~\cite{cvbench}. Following the official evaluation protocols,
we report accuracy for multiple-choice tasks and Mean Relative Accuracy
(MRA) for numerical questions in VSI-Bench. During evaluation on VSI-Bench, we uniformly sample 32 frames from each video for all GUIDE variants.

\noindent\textbf{Training Data.}
We consider a controlled setting and a scaled-data setting. For the
controlled setting, we implement a Qwen3-VL-based reproduction of
VG-LLM~\cite{vgllm}, which fuses the VGGT feature with the visual
tokens only at the LLM input. Following the data configuration of
VG-LLM, we train both this baseline and GUIDE-5B on the same mixture
of SPAR-7M~\cite{spar} and the LLaVA-Hound split of
LLaVA-Video-178K~\cite{llavanext}. The two models use the same MLLM
backbone, training data, frame-sampling strategy,
and optimization settings, enabling a controlled comparison between
input-level fusion and progressive multi-level injection.



For the scaled-data setting, we augment the controlled training
mixture with the training data released by VLM-3R~\cite{vlm3r} and
VSI-590K~\cite{cambrians}. GUIDE-5B$^\dagger$ and
GUIDE-9B$^\dagger$ are trained on this expanded mixture, with 8 frames uniformly sampled per scene. The symbol $^\dagger$
denotes models trained using this scaled-data setting.

\noindent\textbf{Results on VSI-Bench.}
Table~\ref{tab:vsibench_results} reports the results on VSI-Bench.
Under the controlled setting, GUIDE-5B achieves an average score of
55.6, outperforming the VG-LLM-5B$_{\text{Qwen3-VL-4B}}$ baseline by 3.9
points (55.6 vs.\ 51.7). Since the two models use the same backbone,
training data, and configuration, this result provides a
controlled evaluation of the proposed progressive geometric
integration framework.


Under the scaled-data setting, GUIDE-5B$^\dagger$ achieves an average
score of 71.5, improving its controlled-setting counterpart by 15.9
points. Scaling the model from 5B to 9B further raises the score to
72.1. Notably, even GUIDE-5B$^\dagger$ surpasses the strongest
SenseNova-SI-8B variant, trained on over 8 million spatial samples, by
2.7 points (71.5 vs.\ 68.8), while GUIDE-9B$^\dagger$ widens the
margin to 3.3 points. These results demonstrate that GUIDE remains
effective across model scales and benefits from expanded training
data.

To evaluate generalization across diverse spatial reasoning formats, we report comprehensive results in Table~\ref{tab:benchmark_results}. Notably, GUIDE-9B$^\dagger$ achieves the \textbf{highest average score (58.8)}, successfully surpassing even heavyweight proprietary models like GPT-5. Specifically, it achieves the best results on \textbf{VSI-Bench (72.1)} and \textbf{ViewSpatial (48.6)}, while maintaining highly competitive performance on \textbf{CV-Bench (84.9)} and \textbf{MMSI-Bench (29.4)}. These consistent cross-benchmark results verify that our progressive geometric unrolling transfers robustly, proving GUIDE to be a highly generalizable spatial intelligence framework rather than merely overfitting to a single domain.

\begin{table}[t]
\centering
\resizebox{\columnwidth}{!}{%
\begin{tabular}{l cc c}
\toprule
\multirow{2}{*}{\textbf{Model}}
& \multicolumn{2}{c}{\textbf{ScanRefer}}
& \multicolumn{1}{c}{\textbf{Scan2Cap}} \\
\cmidrule(lr){2-3}\cmidrule(lr){4-4}
& \textbf{Acc@0.25}$\uparrow$
& \textbf{Acc@0.5}$\uparrow$
& \textbf{C@0.5}$\uparrow$ \\
\midrule
\rowcolor{gray!10}
\multicolumn{4}{l}{\textit{\textbf{Methods with Explicit 3D Input}}} \\
\shortstack[l]{ScanRefer\\\cite{scanrefer}}& 37.3 & 24.3 & -- \\
MVT~\cite{mvt}
& 40.8 & 33.3 & -- \\
ViL3DRel~\cite{vil3drel}
& 47.9 & 37.7 & -- \\
3D-LLM~\cite{3dllm}
& 30.3 & -- & -- \\
Scan2Cap~\cite{scan2cap}
& -- & -- & 39.1 \\
3DJCG~\cite{3djcg}
& -- & -- & 49.5 \\
D3Net~\cite{d3net}
& -- & -- & 62.6 \\
Vote2Cap-DETR~\cite{vote2cap}
& -- & -- & 61.8 \\
LL3DA~\cite{ll3da}
& -- & -- & 65.2 \\
Chat-3D v2~\cite{chat3dv2}
& 35.9 & 30.4 & 63.9 \\
Grounded 3D-LLM~\cite{grounded3dllm}
& 47.9 & 44.1 & 70.2 \\
LEO~\cite{leo}
& -- & -- & 72.4 \\
ChatScene~\cite{chat3dv2}
& 55.5 & 50.2 & 77.1 \\
LLaVA-3D~\cite{llava3d}
& 54.1 & 42.4 & 79.2 \\
\shortstack[l]{Video-3D LLM (MC)\\\cite{video3dllm}}
& 57.9 & 51.2 & 80.0 \\
\shortstack[l]{Video-3D LLM (Uniform)\\\cite{video3dllm}}
& \textbf{58.1} & \textbf{51.7}
& \textbf{83.8} \\
\midrule
\rowcolor{gray!10}
\multicolumn{4}{l}{\textit{\textbf{Methods with RGB-Only Input}}} \\
SPAR~\cite{spar}
& 31.9 (48.8) & 12.4 (43.1) & -- \\
VG-LLM-4B~\cite{vgllm}
& 36.4 (53.5) & 11.8 (47.5) & 78.6 \\
VG-LLM-8B~\cite{vgllm}
& 41.6 (57.6) & 14.9 (50.9) & 80.0 \\
\rowcolor{gray!10}
\textbf{GUIDE-9B (Ours)}
& \textbf{47.5 (59.6)}
& \textbf{20.5 (53.1)}
& \textbf{81.2} \\
\bottomrule
\end{tabular}%
}
\caption{Results on ScanRefer and Scan2Cap. The best result within
each input setting is highlighted in bold.}
\label{tab:3d_scene_results}
\end{table}

\noindent\subsection{3D Scene Understanding}
\label{sec:3d_perception}

\noindent\textbf{Benchmarks and Evaluation Metrics.}
We evaluate GUIDE on 3D visual grounding using
ScanRefer~\cite{scanrefer} and 3D dense captioning using
Scan2Cap~\cite{scan2cap}. Following SPAR and
VG-LLM, we formulate ScanRefer as video grounding, where
the model predicts the target bounding box in the camera coordinate
system together with its frame index. We report Acc@0.25 and Acc@0.5
at IoU thresholds of 0.25 and 0.5, respectively. For Scan2Cap, the
model generates object descriptions conditioned on object-center
coordinates transformed into the coordinate system of the first frame.
We report CIDEr at IoU 0.5, denoted as C@0.5.

\noindent\textbf{Training Data.}
The training data comprise ScanRefer, Scan2Cap, and the video-based 3D
object detection data introduced by VG-LLM.


\noindent\textbf{Results.}
Table~\ref{tab:3d_scene_results} summarizes the results on ScanRefer
and Scan2Cap. On ScanRefer, the raw predictions of GUIDE-9B achieve
47.5 and 20.5 on Acc@0.25 and Acc@0.5, outperforming VG-LLM-8B by
5.9 and 5.6 points, respectively. These results are obtained using
RGB sequences without explicit 3D inputs. For comparison with prior
work, we also report proposal-refined results in parentheses, obtained
by matching each predicted box to the Mask3D~\cite{mask3d} proposal
with the highest IoU. This refinement increases the scores to 59.6
and 53.1.

On Scan2Cap, GUIDE-9B achieves 81.2 on C@0.5, outperforming
VG-LLM-8B by 1.2 points. Together, the grounding and captioning
results demonstrate the applicability of GUIDE to different 3D scene
understanding tasks.

\begin{table}[t]
\centering
\small
\begin{tabular}{cccc}
\toprule
\textbf{Injection Layers}
& $\boldsymbol{G}_{\mathrm{\mathbf{tok}}}$
& $\boldsymbol{G}_{\mathrm{\mathbf{layer}}}$
& \textbf{Avg.} \\
\midrule

0
& $\times$
& $\times$
& 51.7 \\

\midrule
\multirow{3}{*}{3}
& $\times$     & $\times$     & 52.4 \\
& $\checkmark$ & $\times$     & 52.5 \\
& $\checkmark$ & $\checkmark$ & 53.8 \\

\midrule
\multirow{3}{*}{6}
& $\times$     & $\times$     & 50.3 \\
& $\checkmark$ & $\times$     & 52.7 \\
\rowcolor{gray!10}
& $\checkmark$ & $\checkmark$ & \textbf{55.6} \\

\midrule
\multirow{3}{*}{9}
& $\times$     & $\times$     & 48.0 \\
& $\checkmark$ & $\times$     & 50.4 \\
& $\checkmark$ & $\checkmark$ & 51.1 \\

\midrule
\multirow{3}{*}{24}
& $\times$     & $\times$     & 47.9 \\
& $\checkmark$ & $\times$     & 49.8 \\
& $\checkmark$ & $\checkmark$ & 51.6 \\

\bottomrule
\end{tabular}
\caption{
Ablation study of the number of geometric injection layers and gating
components on VSI-Bench. $G_{\mathrm{tok}}$ and
$G_{\mathrm{layer}}$ denote the token-level gate and layer-level gate,
respectively. 
}
\label{tab:comprehensive_ablation}
\end{table}

\begin{table}[t]
\centering
\small
\begin{tabular}{@{}lcc@{}}
\toprule
\textbf{Strategy} & \textbf{Layer Mapping } & \textbf{Avg.} \\
\midrule
Single Deep Layer
& $\{23\}\!\rightarrow\!\{1{:}6\}$ & 51.6 \\

Deep VGGT Only
& $\{18{:}23\}\!\rightarrow\!\{1{:}6\}$ & 52.0 \\

Early VGGT Only
& $\{8{:}13\}\!\rightarrow\!\{1{:}6\}$ & 53.0 \\

Deep LLM Injection
& $\{8{:}3{:}23\}\!\rightarrow\!\{1{:}4{:}21\}$ & 51.7 \\
\midrule

\rowcolor{gray!10}
\textbf{GUIDE}
& $\{8{:}3{:}23\}\!\rightarrow\!\{1{:}6\}$
& \textbf{55.6} \\
\bottomrule
\end{tabular}
\caption{VSI-Bench ablation of VGGT-to-LLM layer mappings.
$a{:}s{:}b$ denotes indices from $a$ to $b$ with stride $s$;
$a{:}b$ uses unit stride.}
\label{tab:injection_strategy}
\end{table}

\begin{figure}[t]
    \centering
    \includegraphics[width=0.7\columnwidth]
    {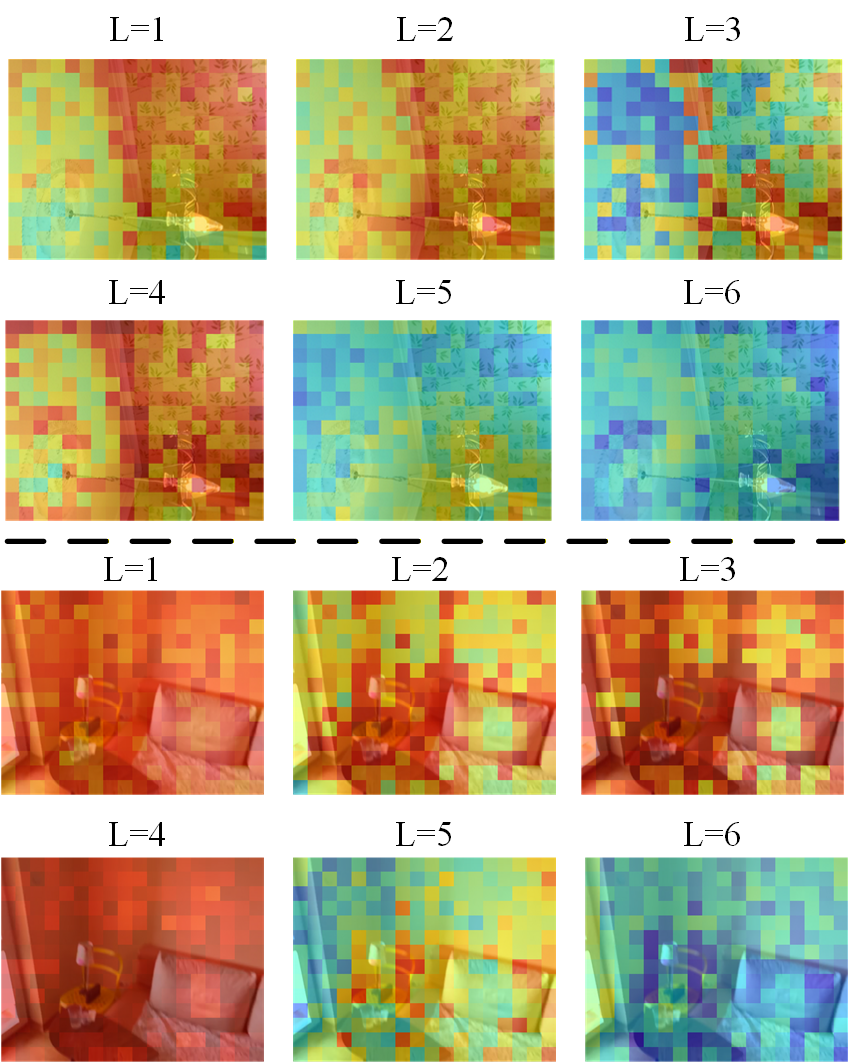}
    \caption{
    Token-level gates across 6 injection layers for two scenes. $L\in\{1,\ldots,6\}$ denotes the index
of the LLM layers. Warmer colors indicate stronger geometric injection
    }
    \label{fig:token_gate_visualization}
\end{figure}

\subsection{Ablation Studies}
\label{sec:ablation}

We conduct component-wise ablations on VSI-Bench using GUIDE-5B under
the controlled training setting. The matched input-fusion baseline,
corresponding to zero injection layers in
Table~\ref{tab:comprehensive_ablation}, adds the 23rd VGGT feature
to the visual tokens only at the LLM input and obtains an average score
of 51.7.

\noindent\textbf{Injection Depth and Dual-Granularity Gating.}
Table~\ref{tab:comprehensive_ablation} jointly studies the number of
injection layers and the token- and layer-level gates. With both gates
enabled, increasing the injection depth from 3 to 6 layers
improves the average score from 53.8 to 55.6. Extending the injection
to 9 or 24 layers instead decreases the scores to 51.1 and 51.6,
indicating that geometric information is most effective when integrated
within a limited number of early LLM layers.

The gating modules consistently improve performance at every injection
depth. For 6-layer injection, direct injection without gating obtains
50.3. Adding the context-aware token-level gate increases the score to
52.7, while further introducing the layer-level global gate improves
it to 55.6. The complete model therefore outperforms ungated 6-layer
injection by 5.3 points and the input-fusion baseline by 3.9 points.
Moreover, ungated injection decreases from 52.4 with 3 layers to
47.9 with 24 layers. These results are consistent with our motivation
that repeated, unregulated injection can introduce redundant geometric
signals and interfere with pretrained representations, while token- and
layer-level regulation mitigates this effect. As visualized in
Figure~\ref{fig:token_gate_visualization}, $G_{\mathrm{tok}}$ exhibits spatially
non-uniform and layer-dependent patterns, supporting its role in
regulating geometric information across visual tokens and
injection layers.

\noindent\textbf{Multi-Level Sampling and Layer Alignment.}
Table~\ref{tab:injection_strategy} examines how geometric features are
selected from VGGT and mapped to the LLM. Repeating the single 23rd-layer
VGGT feature across 6 LLM layers obtains 51.6, whereas using only
consecutive deep or early VGGT features yields 52.0 and 53.0,
respectively. Distributing the same multi-level VGGT features across
deeper LLM layers obtains 51.7. In contrast, GUIDE combines features
sampled across VGGT depths with consecutive early-layer injection and
achieves 55.6. Compared with repeated single-layer features and
distributed deep-LLM injection, GUIDE improves by 4.0 and 3.9 points,
respectively. These comparisons support both the complementary value
of multi-level geometric features and their progressive integration
within early LLM layers.

\section{Conclusion}
\label{sec:conclusion}

We presented \textbf{GUIDE}, a framework for progressively integrating geometric priors into MLLMs. GUIDE samples and aligns multi-level features from a feed-forward geometric encoder and sequentially injects them into early LLM layers, providing evolving visual representations with continuous access to multi-granularity geometric cues. A context-aware token-level gate and a learnable layer-level global gate further regulate this process to reduce redundant injection and limit interference with pretrained representations. Experiments on spatial reasoning and 3D scene understanding benchmarks demonstrate the effectiveness and cross-task applicability of GUIDE using only RGB visual inputs.

\clearpage
\bibliography{main}

@article{geoalign,
  title={GeoAlign: Geometric Feature Realignment for MLLM Spatial Reasoning},
  author={Liu, Zhaochen and Qiao, Limeng and Wan, Guanglu and Jiang, Tingting},
  journal={arXiv preprint arXiv:2604.12630},
  year={2026}
}

@article{gapmllm,
  title={GAP-MLLM: Geometry-Aligned Pre-training for Activating 3D Spatial Perception in Multimodal Large Language Models},
  author={Zhang, Jiaxin and Jiang, Junjun and Li, Haijie and Chen, Youyu and Jiang, Kui and Chen, Dave Zhenyu},
  journal={arXiv preprint arXiv:2603.16461},
  year={2026}
}

@String{Computer = "{IEEE} Computer" }

@String{Springer = "Springer-Verlag" }

@inproceedings{dust3r,
  title={Dust3r: Geometric 3d vision made easy},
  author={Wang, Shuzhe and Leroy, Vincent and Cabon, Yohann and Chidlovskii, Boris and Revaud, Jerome},
  booktitle={Proceedings of the IEEE/CVF conference on computer vision and pattern recognition},
  pages={20697--20709},
  year={2024}
}

@inproceedings{vggt,
  title={Vggt: Visual geometry grounded transformer},
  author={Wang, Jianyuan and Chen, Minghao and Karaev, Nikita and Vedaldi, Andrea and Rupprecht, Christian and Novotny, David},
  booktitle={Proceedings of the Computer Vision and Pattern Recognition Conference},
  pages={5294--5306},
  year={2025}
}

@article{vlm3r,
  title={Vlm-3r: Vision-language models augmented with instruction-aligned 3d reconstruction},
  author={Fan, Zhiwen and Zhang, Jian and Li, Renjie and Zhang, Junge and Chen, Runjin and Hu, Hezhen and Wang, Kevin and Qu, Huaizhi and Wang, Dilin and Yan, Zhicheng and others},
  journal={arXiv preprint arXiv:2505.20279},
  year={2025}
}

@article{spatialmllm,
  title={Spatial-mllm: Boosting mllm capabilities in visual-based spatial intelligence},
  author={Wu, Diankun and Liu, Fangfu and Hung, Yi-Hsin and Duan, Yueqi},
  journal={arXiv preprint arXiv:2505.23747},
  year={2025}
}

@article{spacemind,
  title={SpaceMind: Camera-Guided Modality Fusion for Spatial Reasoning in Vision-Language Models},
  author={Zhao, Ruosen and Zhang, Zhikang and Xu, Jialei and Chang, Jiahao and Chen, Dong and Li, Lingyun and Sun, Weijian and Wei, Zizhuang},
  journal={arXiv preprint arXiv:2511.23075},
  year={2025}
}

@article{vgllm,
  title={Learning from videos for 3d world: Enhancing mllms with 3d vision geometry priors},
  author={Zheng, Duo and Huang, Shijia and Li, Yanyang and Wang, Liwei},
  journal={arXiv preprint arXiv:2505.24625},
  year={2025}
}

@InProceedings{clip,
  title = 	 {Learning Transferable Visual Models From Natural Language Supervision},
  author =       {Radford, Alec and Kim, Jong Wook and Hallacy, Chris and Ramesh, Aditya and Goh, Gabriel and Agarwal, Sandhini and Sastry, Girish and Askell, Amanda and Mishkin, Pamela and Clark, Jack and Krueger, Gretchen and Sutskever, Ilya},
  booktitle = 	 {Proceedings of the 38th International Conference on Machine Learning},
  pages = 	 {8748--8763},
  year = 	 {2021},
  volume = 	 {139},
  series = 	 {Proceedings of Machine Learning Research},
  month = 	 {18--24 Jul}
}

@article{flamingo,
  title={Flamingo: a visual language model for few-shot learning},
  author={Alayrac, Jean-Baptiste and Donahue, Jeff and Luc, Pauline and Miech, Antoine and Barr, Iain and Hasson, Yana and Lenc, Karel and Mensch, Arthur and Millican, Katherine and Reynolds, Malcolm and others},
  journal={Advances in neural information processing systems},
  volume={35},
  pages={23716--23736},
  year={2022}
}

@inproceedings{blip2,
  title={Blip-2: Bootstrapping language-image pre-training with frozen image encoders and large language models},
  author={Li, Junnan and Li, Dongxu and Savarese, Silvio and Hoi, Steven},
  booktitle={International conference on machine learning},
  pages={19730--19742},
  year={2023},
  organization={PMLR}
}

@article{qwen3vl,
  title={Qwen3-vl technical report},
  author={Bai, Shuai and Cai, Yuxuan and Chen, Ruizhe and Chen, Keqin and Chen, Xionghui and Cheng, Zesen and Deng, Lianghao and Ding, Wei and Gao, Chang and Ge, Chunjiang and others},
  journal={arXiv preprint arXiv:2511.21631},
  year={2025}
}

@article{llava,
  title={Visual instruction tuning},
  author={Liu, Haotian and Li, Chunyuan and Wu, Qingyang and Lee, Yong Jae},
  journal={Advances in neural information processing systems},
  volume={36},
  pages={34892--34916},
  year={2023}
}

@inproceedings{internvl,
  title={Internvl: Scaling up vision foundation models and aligning for generic visual-linguistic tasks},
  author={Chen, Zhe and Wu, Jiannan and Wang, Wenhai and Su, Weijie and Chen, Guo and Xing, Sen and Zhong, Muyan and Zhang, Qinglong and Zhu, Xizhou and Lu, Lewei and others},
  booktitle={Proceedings of the IEEE/CVF conference on computer vision and pattern recognition},
  pages={24185--24198},
  year={2024}
}

@inproceedings{vsibench,
  title={Thinking in space: How multimodal large language models see, remember, and recall spaces},
  author={Yang, Jihan and Yang, Shusheng and Gupta, Anjali W and Han, Rilyn and Fei-Fei, Li and Xie, Saining},
  booktitle={Proceedings of the Computer Vision and Pattern Recognition Conference},
  pages={10632--10643},
  year={2025}
}

@inproceedings{ll3da,
  title={Ll3da: Visual interactive instruction tuning for omni-3d understanding reasoning and planning},
  author={Chen, Sijin and Chen, Xin and Zhang, Chi and Li, Mingsheng and Yu, Gang and Fei, Hao and Zhu, Hongyuan and Fan, Jiayuan and Chen, Tao},
  booktitle={Proceedings of the IEEE/CVF conference on computer vision and pattern recognition},
  pages={26428--26438},
  year={2024}
}

@article{gpt4scene,
  title={Gpt4scene: Understand 3d scenes from videos with vision-language models},
  author={Qi, Zhangyang and Zhang, Zhixiong and Fang, Ye and Wang, Jiaqi and Zhao, Hengshuang},
  journal={arXiv preprint arXiv:2501.01428},
  year={2025}
}

@inproceedings{video3dllm,
  title={Video-3d llm: Learning position-aware video representation for 3d scene understanding},
  author={Zheng, Duo and Huang, Shijia and Wang, Liwei},
  booktitle={Proceedings of the IEEE/CVF Conference on Computer Vision and Pattern Recognition},
  pages={8995--9006},
  year={2025}
}

@article{spacer,
  title={SpaceR: Reinforcing MLLMs in Video Spatial Reasoning},
  author={Ouyang, Kun and Liu, Yuanxin and Wu, Haoning and Liu, Yi and Zhou, Hao and Zhou, Jie and Meng, Fandong and Sun, Xu},
  journal={arXiv preprint arXiv:2504.01805},
  year={2025}
}

@inproceedings{
vilasr,
title={Reinforcing Spatial Reasoning in Vision-Language Models with Interwoven Thinking and Visual Drawing},
author={Junfei Wu and Jian Guan and Kaituo Feng and Qiang Liu and Shu Wu and Liang Wang and Wei Wu and Tieniu Tan},
booktitle={The Thirty-ninth Annual Conference on Neural Information Processing Systems},
year={2025},
}

@article{jiang2026spa3r,
  title={Spa3R: Predictive spatial field modeling for 3D visual reasoning},
  author={Jiang, Haoyi and Liu, Liu and Wang, Xinjie and He, Yonghao and Sui, Wei and Su, Zhizhong and Liu, Wenyu and Wang, Xinggang},
  journal={arXiv preprint arXiv:2602.21186},
  year={2026}
}

@inproceedings{
ray2025sat,
title={{SAT}: Dynamic Spatial Aptitude Training for Multimodal Language Models},
author={Arijit Ray and Jiafei Duan and Ellis L Brown II and Reuben Tan and Dina Bashkirova and Rose Hendrix and Kiana Ehsani and Aniruddha Kembhavi and Bryan A. Plummer and Ranjay Krishna and Kuo-Hao Zeng and Kate Saenko},
booktitle={Second Conference on Language Modeling},
year={2025}
}

@inproceedings{
spar,
title={From Flatland to Space: Teaching Vision-Language Models to Perceive and Reason in 3D},
author={Jiahui Zhang and Yurui Chen and Yueming Xu and Ze Huang and Jilin Mei and Junhui Chen and Yanpeng Zhou and Yu-Jie Yuan and Xinyue Cai and Guowei Huang and Xingyue Quan and Hang Xu and Li Zhang},
booktitle={The Thirty-ninth Annual Conference on Neural Information Processing Systems Datasets and Benchmarks Track},
year={2025}
}

@article{cambrians,
  title={Cambrian-s: Towards spatial supersensing in video},
  author={Yang, Shusheng and Yang, Jihan and Huang, Pinzhi and Brown, Ellis and Yang, Zihao and Yu, Yue and Tong, Shengbang and Zheng, Zihan and Xu, Yifan and Wang, Muhan and others},
  journal={arXiv preprint arXiv:2511.04670},
  year={2025}
}

@InProceedings{sensenova-si,
  title = {Scaling Spatial Intelligence with Multimodal Foundation Models},
  author = {Cai, Zhongang and Wang, Ruisi and Gu, Chenyang and Pu, Fanyi and Xu, Junxiang and Wang, Yubo and Yin, Wanqi and Yang, Zhitao and Wei, Chen and Sun, Qingping and Zhou, Tongxi and Li, Jiaqi and Pang, Hui En and Qian, Oscar and Wei, Yukun and Lin, Zhiqian and Shi, Xuanke and Deng, Kewang and Han, Xiaoyang and Chen, Zukai and Fan, Xiangyu and Deng, Hanming and Lu, Lewei and Pan, Liang and Li, Bo and Liu, Ziwei and Wang, Quan and Lin, Dahua and Yang, Lei},
  booktitle = {Proceedings of the IEEE/CVF Conference on Computer Vision and Pattern Recognition (CVPR)},
  year = {2026}
}

@inproceedings{
spatialladder,
title={SpatialLadder: Progressive Training for Spatial Reasoning in Vision-Language Models},
author={Hongxing Li and Dingming Li and Zixuan Wang and Yuchen Yan and Hang Wu and Wenqi Zhang and Yongliang Shen and Weiming Lu and Jun Xiao and Yueting Zhuang},
booktitle={The Fourteenth International Conference on Learning Representations},
year={2026},
}

@inproceedings{mast3r,
  title={Grounding image matching in 3d with mast3r},
  author={Leroy, Vincent and Cabon, Yohann and Revaud, J{\'e}r{\^o}me},
  booktitle={European conference on computer vision},
  pages={71--91},
  year={2024},
  organization={Springer}
}

@inproceedings{scan2cap,
  title={Scan2cap: Context-aware dense captioning in rgb-d scans},
  author={Chen, Zhenyu and Gholami, Ali and Nie{\ss}ner, Matthias and Chang, Angel X},
  booktitle={Proceedings of the IEEE/CVF conference on computer vision and pattern recognition},
  pages={3193--3203},
  year={2021}
}

@inproceedings{3djcg,
  title={3djcg: A unified framework for joint dense captioning and visual grounding on 3d point clouds},
  author={Cai, Daigang and Zhao, Lichen and Zhang, Jing and Sheng, Lu and Xu, Dong},
  booktitle={Proceedings of the IEEE/CVF conference on computer vision and pattern recognition},
  pages={16464--16473},
  year={2022}
}

@inproceedings{d3net,
title = {D3Net: A Unified Speaker-Listener Architecture for 3D Dense Captioning and Visual Grounding},
author = {Chen, Dave Zhenyu and Wu, Qirui and Nie{\ss}ner, Matthias and Chang, Angel X},
year = {2022},
booktitle = {ECCV},
pages = {487--505}
}

@inproceedings{vote2cap,
  title={End-to-end 3d dense captioning with vote2cap-detr},
  author={Chen, Sijin and Zhu, Hongyuan and Chen, Xin and Lei, Yinjie and Yu, Gang and Chen, Tao},
  booktitle={Proceedings of the IEEE/CVF conference on computer vision and pattern recognition},
  pages={11124--11133},
  year={2023}
}

@article{chat3dv2,
  title={Chat-scene: Bridging 3d scene and large language models with object identifiers},
  author={Huang, Haifeng and Chen, Yilun and Wang, Zehan and Huang, Rongjie and Xu, Runsen and Wang, Tai and Liu, Luping and Cheng, Xize and Zhao, Yang and Pang, Jiangmiao and others},
  journal={Advances in Neural Information Processing Systems},
  volume={37},
  pages={113991--114017},
  year={2024}
}

@article{grounded3dllm,
  title={Grounded 3d-llm with referent tokens},
  author={Chen, Yilun and Yang, Shuai and Huang, Haifeng and Wang, Tai and Xu, Runsen and Lyu, Ruiyuan and Lin, Dahua and Pang, Jiangmiao},
  journal={arXiv preprint arXiv:2405.10370},
  year={2024}
}

@inproceedings{
leo,
title={An Embodied Generalist Agent in 3D World},
author={Jiangyong Huang and Silong Yong and Xiaojian Ma and Xiongkun Linghu and Puhao Li and Yan Wang and Qing Li and Song-Chun Zhu and Baoxiong Jia and Siyuan Huang},
booktitle={Forty-first International Conference on Machine Learning},
year={2024},
}

@inproceedings{llava3d,
  title={Llava-3d: A simple yet effective pathway to empowering lmms with 3d capabilities},
  author={Zhu, Chenming and Wang, Tai and Zhang, Wenwei and Pang, Jiangmiao and Liu, Xihui},
  booktitle={Proceedings of the IEEE/CVF International Conference on Computer Vision},
  pages={4295--4305},
  year={2025}
}

@inproceedings{scanrefer,
  title={Scanrefer: 3d object localization in rgb-d scans using natural language},
  author={Chen, Dave Zhenyu and Chang, Angel X and Nie{\ss}ner, Matthias},
  booktitle={European conference on computer vision},
  pages={202--221},
  year={2020},
  organization={Springer}
}

@inproceedings{mvt,
  title={Multi-view transformer for 3d visual grounding},
  author={Huang, Shijia and Chen, Yilun and Jia, Jiaya and Wang, Liwei},
  booktitle={Proceedings of the IEEE/CVF Conference on Computer Vision and Pattern Recognition},
  pages={15524--15533},
  year={2022}
}

@article{vil3drel,
  title={Language conditioned spatial relation reasoning for 3d object grounding},
  author={Chen, Shizhe and Guhur, Pierre-Louis and Tapaswi, Makarand and Schmid, Cordelia and Laptev, Ivan},
  journal={Advances in neural information processing systems},
  volume={35},
  pages={20522--20535},
  year={2022}
}

@article{3dllm,
  title={3d-llm: Injecting the 3d world into large language models},
  author={Hong, Yining and Zhen, Haoyu and Chen, Peihao and Zheng, Shuhong and Du, Yilun and Chen, Zhenfang and Gan, Chuang},
  journal={Advances in Neural Information Processing Systems},
  volume={36},
  pages={20482--20494},
  year={2023}
}

@article{gemini2.5,
  title={Gemini 2.5: Pushing the frontier with advanced reasoning, multimodality, long context, and next generation agentic capabilities},
  author={Comanici, Gheorghe and Bieber, Eric and Schaekermann, Mike and Pasupat, Ice and Sachdeva, Noveen and Dhillon, Inderjit and Blistein, Marcel and Ram, Ori and Zhang, Dan and Rosen, Evan and others},
  journal={arXiv preprint arXiv:2507.06261},
  year={2025}
}

@article{gpt5,
  title={Openai gpt-5 system card},
  author={Singh, Aaditya and Fry, Adam and Perelman, Adam and Tart, Adam and Ganesh, Adi and El-Kishky, Ahmed and McLaughlin, Aidan and Low, Aiden and Ostrow, AJ and Ananthram, Akhila and others},
  journal={arXiv preprint arXiv:2601.03267},
  year={2025}
}

@article{internvl3,
  title={Internvl3: Exploring advanced training and test-time recipes for open-source multimodal models},
  author={Zhu, Jinguo and Wang, Weiyun and Chen, Zhe and Liu, Zhaoyang and Ye, Shenglong and Gu, Lixin and Tian, Hao and Duan, Yuchen and Su, Weijie and Shao, Jie and others},
  journal={arXiv preprint arXiv:2504.10479},
  year={2025}
}

@article{bagel,
  title={Emerging properties in unified multimodal pretraining},
  author={Deng, Chaorui and Zhu, Deyao and Li, Kunchang and Gou, Chenhui and Li, Feng and Wang, Zeyu and Zhong, Shu and Yu, Weihao and Nie, Xiaonan and Song, Ziang and others},
  journal={arXiv preprint arXiv:2505.14683},
  year={2025}
}

@inproceedings{vila1.5,
  title={Vila: On pre-training for visual language models},
  author={Lin, Ji and Yin, Hongxu and Ping, Wei and Molchanov, Pavlo and Shoeybi, Mohammad and Han, Song},
  booktitle={Proceedings of the IEEE/CVF conference on computer vision and pattern recognition},
  pages={26689--26699},
  year={2024}
}

@article{llavanext,
title={Video Instruction Tuning With Synthetic Data},
author={Yuanhan Zhang and Jinming Wu and Wei Li and Bo Li and Zejun Ma and Ziwei Liu and Chunyuan Li},
year={2024},
journal={arXiv preprint arXiv:2410.02713},
}

@article{vst,
  title={Visual spatial tuning},
  author={Yang, Rui and Zhu, Ziyu and Li, Yanwei and Huang, Jingjia and Yan, Shen and Zhou, Siyuan and Liu, Zhe and Li, Xiangtai and Li, Shuangye and Wang, Wenqian and others},
  journal={arXiv preprint arXiv:2511.05491},
  year={2025}
}

@article{mmsibench,
  title={Mmsi-bench: A benchmark for multi-image spatial intelligence},
  author={Yang, Sihan and Xu, Runsen and Xie, Yiman and Yang, Sizhe and Li, Mo and Lin, Jingli and Zhu, Chenming and Chen, Xiaochen and Duan, Haodong and Yue, Xiangyu and others},
  journal={arXiv preprint arXiv:2505.23764},
  year={2025}
}

@article{viewspatial,
  title={Viewspatial-bench: Evaluating multi-perspective spatial localization in vision-language models},
  author={Li, Dingming and Li, Hongxing and Wang, Zixuan and Yan, Yuchen and Zhang, Hang and Chen, Siqi and Hou, Guiyang and Jiang, Shengpei and Zhang, Wenqi and Shen, Yongliang and others},
  journal={arXiv preprint arXiv:2505.21500},
  year={2025}
}

@article{cvbench,
  title={CVBench: Benchmarking Cross-Video Synergies for Complex Multimodal Reasoning},
  author={Zhu, Nannan and Dong, Yonghao and Wang, Teng and Li, Xueqian and Deng, Shengjun and Wang, Yijia and Hong, Zheng and Geng, Tiantian and Niu, Guo and Huang, Hanyan and others},
  journal={arXiv preprint arXiv:2508.19542},
  year={2025}
}

@inproceedings{mask3d,
  title={Mask3d: Mask transformer for 3d semantic instance segmentation},
  author={Schult, Jonas and Engelmann, Francis and Hermans, Alexander and Litany, Or and Tang, Siyu and Leibe, Bastian},
  booktitle={2023 IEEE International Conference on Robotics and Automation (ICRA)},
  pages={8216--8223},
  year={2023},
  organization={IEEE}
}

@inproceedings{align,
  title={Scaling up visual and vision-language representation learning with noisy text supervision},
  author={Jia, Chao and Yang, Yinfei and Xia, Ye and Chen, Yi-Ting and Parekh, Zarana and Pham, Hieu and Le, Quoc and Sung, Yun-Hsuan and Li, Zhen and Duerig, Tom},
  booktitle={International conference on machine learning},
  pages={4904--4916},
  year={2021},
  organization={PMLR}
}


\end{document}